\RequirePackage{fix-cm}
\documentclass[twocolumn]{svjour3}          % twocolumn
\smartqed  % flush right qed marks, e.g. at end of proof
\usepackage{graphicx}
\usepackage{times}
\usepackage{latexsym}
\usepackage[table,xcdraw]{xcolor}
\usepackage{array, makecell} %
\usepackage{xcolor}
\usepackage{siunitx}

\usepackage{microtype}

\begin{document}

\title{Do We Need Online NLU Tools?%\thanks{Grants or other notes
%about the article that should go on the front page should be
%placed here. General acknowledgments should be placed at the end of the article.}
}
\subtitle{Benchmark of Public Intent Recognition Services}

%\titlerunning{Short form of title}        % if too long for running head

\author{Petr Lorenc         \and
        Petr Marek          \and
        Jan Pichl          \and
        Jakub Konrád          \and
        Jan Šedivý      
}

%\authorrunning{Short form of author list} % if too long for running head

\institute{Petr Lorenc,  Petr Marek, Jan Pichl, Jakub Konrád \at
              Faculty of Electrical Engineering, CTU Prague \\
              \email{\{lorenpe2, marekp17, pichljan, konrajak\}@fel.cvut.cz}           %  \\
           \and
           Jan Šedivý \at
              CIIRC, CTU Prague \\
              jan.sedivy@cvut.cz
}

\date{Received: date / Accepted: date}
% The correct dates will be entered by the editor

\maketitle

\begin{abstract}
The intent recognition is an essential algorithm of any conversational AI application. It is responsible for the classification of an input message into meaningful classes. In many bot development platforms, we can configure the NLU pipeline. Several intent recognition services are currently available as an API, or we choose from many open-source alternatives. However, there is no comparison of intent recognition services and open-source algorithms. Many factors make the selection of the right approach to intent recognition challenging in practice. In this paper, we suggest criteria to choose the best intent recognition algorithm for an application. We present a novel CIIRC dataset that evaluates the impact of noise on intent recognition algorithms. We selected types of noise representing typical real-world scenarios. Finally, we compare the selected public NLU services with selected open-source algorithms for intent recognition using the suggested criteria.
\end{abstract}

\section{Introduction}
As a specific field of text classification, intent recognition is a core component of dialogue systems \cite{Pichl2018}. 
There are two main sets of algorithms for intent recognition, unsupervised and supervised. Unsupervised algorithms are the simplest. They try to find the semantically closest one from $m$ predefined sentences grouped into $n$ classes, each class representing one intent. The Semantic Text Similarity(STS) algorithms typically use sentence and query distributions. Both distributions are calculated as a plain or weighted average of the word distributions. The word distributions are created by word2vec \cite{DBLP:journals/corr/MikolovSCCD13_word2vec} or fastText \cite{DBLP:journals/corr/BojanowskiGJM16_fasttext} algorithms. However, the average-based distributions do not take the order of words into account. The latest systems based on recurrent neural networks \cite{hochreiter1997long_LSTM} or self-attention \cite{attentionis2017} aim at the order of words and the words' relative importance. They generate more accurate sentences and query distributions. The cosine similarity is the most frequently used similarity measure.

The intent recognition can also be considered as a classification task in its essence. The supervised algorithms use the same sentence distributions as unsupervised algorithms, but instead of applying cosine similarity, we train the models to classify the query to one of the $n$ classes. The classifiers come in many incarnations. We can choose starting from the logistic regression up to neural networks. Compared to STS algorithms, the supervised models need training. Training takes some time, which is slowing the process of conversational application development. On the other hand, the supervised algorithms are generally achieving higher accuracy.

With the growing complexity of models, it is challenging to choose the best algorithm for a particular task. Moreover, the software developers are overwhelmed by NLU services as well as many open-source algorithms. Public NLU services are usually offered as black boxes that are undergoing development in time. The development progress in the open-source domain is also dramatic. Therefore, trying to find a process for a systematic evaluation of public NLU services and their comparison with open-source algorithms is well justified. 

First, we need to choose the parameters for quality evaluation. The most important parameters are accuracy and recall, or we may use the F1 score. Another parameter is the speed of classification during the runtime and the training duration. The time to classify is important for sizing the server for particular peak traffic. It is also valuable to estimate the required memory size.

We will test public NLU services and open-source algorithms on the same testing sets for evaluation. Selecting the testing set is a nontrivial task. The Conversational AI dialogues may differ significantly. Therefore, it is important to evaluate the algorithms on datasets closely related to the application. The testing set must consider the users' behavior. For evaluation, we try to prepare data sets reflecting an average conversation. We use as a basis the datasets presented in \cite{Braun2018}. We include the most typical errors like misspelling~\cite{Hagiwara2019_github_typos,Edizel2019_misspeling}, filler words, or header phrases, and tail phrases \cite{Shivakumar2019,Pichl2018}. We want to model the hesitation, uncertainty, and staggering. We do not consider the automatic speech recognition (ASR) errors in our study. Tests on such data lead to a comprehensive comparison that we can utilize in real-world scenarios.

We provide the code\footnote{https://github.com/petrLorenc/benchmark-intent-tools} for testing and comparison of public NLU services with open-source algorithms for intent recognition. We selected a set of the most popular public NLU services for evaluation that were also the objects of previous scientific evaluations \cite{Braun2018,Liu2019_benchmark_NLU}. The selected public NLU services are:
\begin{itemize} 
    \item Dialogflow\footnote{https://cloud.google.com/dialogflow}
    \item LUIS\footnote{https://www.luis.ai/}
    \item IBM Watson Assistant\footnote{https://www.ibm.com/cloud/watson-assistant/}
\end{itemize}

We also selected the following open-source algorithms:
\begin{itemize}
    \item fastText \cite{DBLP:journals/corr/BojanowskiGJM16_fasttext}
    \item Sent2Vec \cite{pgj2017sent2vec}
    \item Sentence-BERT \cite{Reimers2019}
    \item Universal Sentence Encoder \cite{Cer2018}
    \item Rasa\footnote{https://rasa.com/} \cite{bocklisch2017rasa}
\end{itemize}

\section{Related work}
Many of the text classification algorithms are taking advantage of the word2vec \cite{DBLP:journals/corr/MikolovSCCD13_word2vec} and fastText \cite{DBLP:journals/corr/BojanowskiGJM16_fasttext} embeddings. The distribution of embedding vector spaces preserves the semantic relations. Therefore, a set of unsupervised algorithms takes advantage of preserved semantic relations for classification based on similarity measurement \cite{focil2017tweets,barros2019unsupervised,dai2017social}. Using the average of word embeddings is the standard practice \cite{Pichl2018} for obtaining the sentence distribution. Such algorithms do not model the order of the words in a sentence. Thus, they are limited to tasks working with shorter and simpler sentences.

For complicated tasks, the recurrent neural networks, like LSTM \cite{hochreiter1997long_LSTM} are generating sentence distributions preserving the order of words. The recent attention-based models achieve even better accuracy \cite{attentionis2017}.  One of the most popular algorithms is the general attention-based model BERT \cite{BERT2019} or specialized models like Universal Sentence Encoder (USE) \cite{Cer2018} or Sentence-BERT \cite{Reimers2019}. These algorithms are coming with pretrained models designed for transfer learning. With a reasonably small number of computations, we can train good intent classifiers. There are also works proposing efficiency improvements of such models \cite{FastBert2020}.

The real-world intent recognition encounters several sources of noise. The work of Shivakumar et al. \cite{Shivakumar2019} describes how Automatic Speech Recognition adds noise to the input data. Hagiwara et al. \cite{Hagiwara2019_github_typos} show the effects of misspelling when the user types the message. Older approaches based on word embedding can have a problem dealing with this noise \cite{Edizel2019_misspeling}. Unfortunately, Sun et al. \cite{Sun2020_typos_bert} demonstrate that even the state-of-the-art models do not solve these problems altogether.

Generative adversarial networks \cite{Salvaris_2018_GAN}, which are usually used for image classification, can also generate training examples in the text domain. Liang et al. \cite{DBLP:journals/corr/LiangLSBLS17_GAN_fooled_neural_network} also shows how adversarially generated examples fools neural network models.

Only a few papers focus on a comparison between public NLU services and open-source algorithms for intent recognition. Braun et al.~\cite{Braun2018} started to explicitly discuss why the software architect should choose one NLU service over another and how this decision may influence their system's performance. The evaluation presented in the paper reflects a common situation when we do not have enough training data. Liu et al. \cite{Liu2019_benchmark_NLU} benchmarked the online services and provided the data for 10-fold cross-validation publicly. However, both of these works lack the comparison with the open-source algorithms.

\section{Experiments}

\begin{table*}[ht]
\centering
\caption{Abbreviations for datasets and text transformations}
\resizebox{2\columnwidth}{!}{%
\begin{tabular}{ll|l} \hline
\textbf{Nameset}      & \textbf{}                       & \textbf{Description}                                                              \\ \hline
\textbf{Original}     &                                 &                                                                      \\
&       \textbf{Chatbot Corpus - Chat}                          & Questions from a Telegram chatbot about public transport \\
&       \textbf{Ask Ubuntu Corpus - Ask}                        & Questions and answers from https://askubuntu.com \\
&       \textbf{Web Applications Corpus - Web}                        &  Questions and answers from https://webapps.stackexchange.com\\ \hline
&       \textbf{Enrich}                                 & Added Header and Footer phrases, random word in sentence is replaced by filler word \\
&       \textbf{Typos}                                  & Swap a character in all sentences               \\
&       \textbf{Typos 50\%}                               & Swap a character in 50\% of sentences                \\
&       \textbf{Filler words}                           & Random word in sentence is replaced by filler word                                 \\ \hline
\textbf{CIIRC Small}&                                    &                                              \\
\textbf{}             & \textbf{HF}                     & Added Header and Footer phrases                                                   \\
\textbf{}             & \textbf{HFR}                    & Added Header and Footer phrases, Replace filler word with random word in sentence \\
\textbf{CIIRC}       &                                 &                                           \\
\textbf{}             & \textbf{HF}                     & Added Header and Footer phrases                                                   \\
\textbf{}             & \textbf{HFR}                    & Added Header and Footer phrases, random word in sentence is replaced by filler word \\  
\textbf{}             & \textbf{HFR-Tr}                 & Added Header and Footer phrases, random word in training sentences is replaced by filler word \\  
\textbf{}             & \textbf{HFR-Te}                 & Added Header and Footer phrases, random word in testing sentences is replaced by filler word \\  \hline
                          
\end{tabular}%
}
\label{tab:description_data}
\end{table*}

\subsection{Datasets}
We used the datasets presented in \cite{Braun2018} as our starting point. There are Chatbot Corpus (Chat), Ask Ubuntu Corpus (Ask), and Web Applications Corpus (Web). The Chat dataset is based on questions gathered by a Telegram chatbot in production, answering questions about public transport connections. The Ask and Web corpora are based on data from StackExchange platforms called Ask Ubuntu, and Web Applications. The datasets consist of the most popular questions asked on those platforms. The questions were annotated by intent using Amazon Mechanical Turk. Table \ref{tab:braun_data} shows the datasets' class distribution. The selected datasets evaluate intent classifiers' ability to learn from a small number of examples in three different domains.

As the datasets presented in \cite{Braun2018} contain tiny amounts of examples per each intent class and do not fully represent the data distribution for conversational AI applications, we also prepared and evaluated algorithms on the CIIRC dataset, which consists of intent examples that we created for a word game application. It includes 14 intents: ``Yes'', ``No'', ``Maybe'', ``Repeat'', ``Wait'', ``Play Other Game'', ``List Of Games'', ``Games Counter'', ``Volume Up'', ``Volume Down'', ``Time Played'', ``Still There'', ``Stop'', and ``Total Stop''. We also created a version of the dataset consisting of a limited number of examples. We call it CIIRC Small. The CIIRC dataset contains 659 training and 660 testing examples. The CIIRC Small dataset contains 154 training and 220 testing examples. The data distribution of both datasets represents the typical distribution of training datasets for conversational AI applications.

In real-world scenarios, applications' users introduce noise into the input. To evaluate the impact of noise on the algorithms, we selected the types of noise to match the typical real-world user behavior described in Pichl et al.~\cite{Pichl2018} and observed during the development of socialbot Alquist~\cite{pichl2018alquist,pichl2020alquist} for Amazon Alexa Prize Competitions~\cite{ram2018conversational,khatri2018alexa,gabriel2020further}. The types of noise can be divided into noise typical for voice applications and noise typical for text applications.

The typical noise for voice applications is caused by flaws in ASR and indecisive users. We simulate these flaws by adding filler words into the sentence, prepending header phrases, and appending footer phrases. We present the list of filler words, header, and footer phrases in Table \ref{tab:noise_types}.

% Please add the following required packages to your document preamble:
% \usepackage{graphicx}
\begin{table}
\caption{Filler words, header and footer phrases.}
\label{tab:noise_types}
\resizebox{\columnwidth}{!}{%
\begin{tabular}{l|l}
\hline
\textbf{\begin{tabular}[c]{@{}l@{}}Filler\\ Words\end{tabular}}   & yeah, oh, ehm, uhm, just, really, somehow                                                                                                                                                                                                                                                                     \\ \hline
\textbf{\begin{tabular}[c]{@{}l@{}}Header\\ Phrases\end{tabular}} & \begin{tabular}[c]{@{}l@{}}I think, I wonder, you know, sounds great,\\ okay, so, right, believe me, at the end of\\ the day, basically, well, would you let me\\ know, I would like to hear, I want you to,\\ could you, can you, in my humble opinion,\\ for what it is worth, needless to say\end{tabular} \\ \hline
\textbf{\begin{tabular}[c]{@{}l@{}}Footer\\ Phrases\end{tabular}} & \begin{tabular}[c]{@{}l@{}}please, right, you know what I mean, you \\ see, seriously, as you can see, you know, well\end{tabular}                                                                                                                                                                            \\ \hline
\end{tabular}%
}
\end{table}

The noise typical for text applications are typos. We simulate it by randomly changing the position of nearby characters.

Table \ref{tab:description_data} presents abbreviations for all datasets we used for evaluation and shows the detailed description of the transformations we used to introduce the noise.

\begin{table}
\centering
\caption{Number of examples per intent class in Chat, Ask and Web datasets.}
\resizebox{0.9\columnwidth}{!}{%
\begin{tabular}{l|l|cc}
\hline
\textbf{Dataset}                 & \textbf{Intent}                  & \textbf{Train} & \textbf{Test} \\ \hline
\textbf{Chatbot}           & FindConnection          & 57    & 71   \\
\textbf{Corpus}                        & DepartureTime           & 43    & 35   \\
                        & TOTAL           & 100    & 106   \\ \hline
\textbf{Ask Ubuntu}         & None                    & 3     & 5    \\
\textbf{Corpus}                        & Make Update             & 10    & 37   \\
                        & Setup Printer           & 10    & 13   \\
                        & Software Recommendation & 17    & 40   \\
                        & Shutdown Computer       & 13    & 14   \\
                        & TOTAL       & 66    & 123   \\\hline
\textbf{Web Applications} & Download Video          & 1     & 0    \\
\textbf{Corpus}                        & Change Password         & 2     & 6    \\
                        & None                    & 2     & 4    \\
                        & Export Data             & 2     & 3    \\
                        & Sync Accounts           & 3     & 6    \\
                        & Filter Spam             & 6     & 14   \\
                        & Find Alternatives       & 7     & 16   \\
                        & Delete Account          & 7     & 10   \\
                        & TOTAL          & 30     & 59   \\
                        \hline
\end{tabular}%
}
\label{tab:braun_data}
\end{table}

\subsection{Metrics}
We based our evaluation of public NLU services and open-source algorithms on four criteria: classification accuracy, robustness towards noise, computational requirements, and cost. We selected these criteria as they represent the crucial properties of any classification algorithm.

We use the F1 score to evaluate the classification accuracy. We calculate the precision and recall for the F1 formula using the sum of the individual true positives, false positives, and false negatives of the system for different intent classes as described by Opitz et al. \cite{macro_f1_score}.

To evaluate a robustness towards noise, we calculate the F1 score on the dataset into which we introduced noise. We added header and footer phrases, filler words, and typos. Datasets enhanced by those types of noise evaluate the real-world performance better.

To estimate the computational requirements, we measure how much RAM and how many CPU cores each algorithm requires. We use the {\fontfamily{cmtt}\selectfont free} command to measure memory requirements and the {\fontfamily{cmtt}\selectfont top} command to measure CPU core utilization. We report only the number of CPU cores since it defines the size of the virtual machine.

We instantiated the cheapest AWS EC2 instance with the algorithm's required memory and CPU cores to evaluate the cost. The Table \ref{tab:ec2} shows the EC2 parameters we used in our experiments. We calculate the price per one request in USD by dividing the hourly cost of an instance by the number of requests the algorithm handles in one hour. Our selected evaluation is an approximation of the peak traffic the conversational AI application could handle. We do not normalize the cost for the same number of requests per hour, which is a significant simplification, but it is indicative enough for an algorithm selection. We considered only AWS because cloud computational services are highly competitive, and alternative services like Azure or Google Cloud provide comparative solutions and prices. Thus, AWS serves as a representative for those services. For public NLU services, we took the price from official documentation and converted it to the price for one request in USD.

% Please add the following required packages to your document preamble:
% \usepackage{graphicx}
\begin{table}
\caption{Parameters of used EC2 instances}
\label{tab:ec2}
\resizebox{\columnwidth}{!}{%
\begin{tabular}{l|l}
\hline
\textbf{EC2 Instance} & \textbf{Parameters}                                                                                  \\ \hline
\textbf{t4g.micro}    & \begin{tabular}[c]{@{}l@{}}vCPUs: 2\\ Memory: 4 GiB\\ On-Demand Hourly Cost: 0.0084 USD\end{tabular} \\ \hline
\textbf{a1.xlarge}    & \begin{tabular}[c]{@{}l@{}}vCPUs: 4\\ Memory: 8 GiB\\ On-Demand Hourly Cost: 0.102 USD\end{tabular}  \\ \hline
\textbf{a1.2xlarge}   & \begin{tabular}[c]{@{}l@{}}vCPUs: 8\\ Memory: 16 GiB\\ On-Demand Hourly Cost: 0.204 USD\end{tabular} \\ \hline
\end{tabular}%
}
\end{table}

% Please add the following required packages to your document preamble:
% \usepackage{graphicx}
\begin{table*}[ht]
\centering
\caption{F1 score comparison of fastText and Sent2Vec embeddings using cosine similarity and logistic regression. We compare 300-dimensional models trained on the same dataset and 300-dimensional and 700-dimensional pretrained publicly available models.}
\label{tab:embeddings}
\resizebox{2\columnwidth}{!}{%
\begin{tabular}{l|cc|cc|cc|cc}
\hline
                      & \multicolumn{4}{c|}{\textbf{Cosine Similarity}}                                                                                                                           & \multicolumn{4}{c}{\textbf{Logistic Regression}}                                                                                                                          \\ \hline
\textbf{}             & \multicolumn{2}{c|}{\textbf{fastText}}                                                & \multicolumn{2}{c|}{\textbf{Sent2Vec}}                                             & \multicolumn{2}{c|}{\textbf{fastText}}                                               & \multicolumn{2}{c}{\textbf{Sent2Vec}}                                               \\
\textbf{Corpus}       & \textbf{300D}  & \textbf{\begin{tabular}[c]{@{}c@{}}300D \\ pretrained\end{tabular}} & \textbf{300D} & \textbf{\begin{tabular}[c]{@{}c@{}}700D\\ pretrained\end{tabular}} & \textbf{300D} & \textbf{\begin{tabular}[c]{@{}c@{}}300D \\ pretrained\end{tabular}} & \textbf{300D}  & \textbf{\begin{tabular}[c]{@{}c@{}}700D \\ pretrained\end{tabular}} \\ \hline
\textbf{Ask}          & 0,568          & 0,678                                                               & 0,715         & 0,807                                                              & 0,798         & 0,798                                                               & 0,834          & \textbf{0,908}                                                      \\
\textbf{Chat}         & \textbf{0,981} & 0,972                                                               & 0,971         & 0,933                                                              & 0,971         & \textbf{0,981}                                                      & 0,943          & 0,971                                                               \\
\textbf{Web}          & 0,559          & 0,780                                                               & 0,644         & 0,745                                                              & 0,627         & \textbf{0,847}                                                      & 0,762          & \textbf{0,847}                                                      \\ \hline
\textbf{CIIRC Small}  & 0,786          & 0,827                                                               & 0,800         & 0,891                                                              & 0,877         & 0,882                                                               & 0,881          & \textbf{0,895}                                                      \\
\textbf{CIIRC}        & 0,910          & 0,905                                                               & 0,927         & 0,947                                                              & 0,943         & 0,929                                                               & \textbf{0,953} & \textbf{0,953}                                                      \\
\textbf{CIIRC-HFR-Tr} & 0,843          & 0,864                                                               & 0,868         & 0,912                                                              & 0,943         & 0,945                                                               & 0,942          & \textbf{0,958}                                                      \\
\textbf{CIIRC-HFR-Te} & 0,722          & 0,805                                                               & 0,875         & \textbf{0,914}                                                     & 0,865         & 0,859                                                               & 0,880          & 0,912                                                               \\ \hline
\end{tabular}%
}
\end{table*}

\subsection{Methods}
We selected Dialogflow, LUIS, and IBM Watson Assistant. Works of Braun et al. \cite{Braun2018} and Liu et al. \cite{Liu2019_benchmark_NLU} compared those public NLU services as well. However, those comparisons were conducted in 2018 and 2019, and those services underwent further development.

\begin{itemize}
    \item \emph{Dialogflow}\\We disabled beta features, set the classification threshold to 0, selected ML mode, and disabled spell checking. We used the default values for the rest of the settings.\\
    
    \item \emph{LUIS}\\We used non-deterministic training. We enabled normalizing punctuation and word forms. We used the default values for the rest of the settings.\\
    
    \item \emph{IBM Watson Assistant}\\We disabled disambiguation and autocorrection, and we set irrelevance detection to \emph{Keep existing}. We used the default values for the rest of the settings.
\end{itemize}

We selected the following open-source algorithms as viable alternatives to public NLU services for conversational AI applications:

\begin{itemize} 
    \item Word embeddings avg. + Cosine similarity, SVM or Logistic regression
    \item Universal Sentence Encoder \cite{Cer2018} + Cosine similarity, SVM or Logistic regression
    \item Sentence-BERT \cite{Reimers2019} + Cosine similarity, SVM or Logistic regression
    \item Rasa 2.0 \cite{bocklisch2017rasa}
\end{itemize}

For the supervised approach, we have chosen the latest attention-based open-source algorithms. They deliver the state-of-the-art performance for many NLP tasks. They come with models trained on large databases in the order of TB. Transfer learning training is an ideal solution to utilize the power of large models. Since the number of example intents is small, we have used the standard classification algorithms logistic regression \cite{mccullagh2018generalized} and SVM \cite{cortes1995support}.

\begin{itemize}
    \item \emph{Word embeddings}\\We assume fastText \cite{DBLP:journals/corr/BojanowskiGJM16_fasttext} and Sent2Vec \cite{pgj2017sent2vec}. We use Sent2Vec vectors pretrained on Twitter\footnote{https://github.com/epfml/sent2vec\#downloading-sent2vec-pre-trained-models} and fastText vectors pretrained on Wikipedia\footnote{https://fasttext.cc/docs/en/crawl-vectors.html}. We perform vector compression for both models. Works of \cite{fastextZip} and \cite{Brich2018} show that compression does not decrease the performance significantly and leads to a considerable reduction of the memory footprint of less than 15MB. The sentence distribution is computed by averaging the embedding vector of each word in the sentence.\\
    
    \item \emph{Logistic regression}\\We used scikit-learn's \cite{scikit-learn} implementation of logistic regression. We used default parameters for logistic regression except for the parameters \emph{inversion of the regularization strength} that we set to 1000 and the \emph{number of max iterations} that we set to 1000.\\
    
    \item \emph{Support Vector Machine}\\We used scikit-learn's \cite{scikit-learn} implementation of the support vector machine. We used the default parameters. We only set the parameter \emph{probability} to true to receive a probability estimate of the classification.\\
    
    \item \emph{Sentence-BERT}\\We used weights from \emph{bert-base-nli-stsb-mean-tokens}\footnote{https://huggingface.co/sentence-transformers/bert-base-nli-stsb-mean-tokens} for Sentence-BERT.\\
    
    \item \emph{Universal Sentence Encoder}\\We used Universal Sentence Encoder version 4\footnote{Model can be downloaded from https://tfhub.dev/google/universal-sentence-encoder/4}.\\
    
    \item \emph{Rasa}\\Rasa uses DIET \cite{bunk2020diet} architecture for intent and named entity recognition. We used the model's intent recognition part and executed it with default parameters\footnote{https://github.com/RasaHQ/rasa/blob/master/data/test\_config/config\_\\defaults.yml}. 
\end{itemize}

\subsection{Results}
Firstly, we conducted two experiments evaluating the impact of fastText and Sent2Vec embeddings on open-source algorithms. The first experiment tested the performance of publicly available pretrained models. We compared the 300-dimensional pretrained English fastText vectors trained on Wikipedia\footnote{https://fasttext.cc/docs/en/pretrained-vectors.html} with 700-dimensional Sent2Vec unigram vectors pretrained on Twitter\footnote{https://github.com/epfml/sent2vec}.

We conducted experiments using two embedding models with a different number of dimensions for two reasons. We believe that the authors of pretrained models selected the proper size of the vector space \cite{DBLP:journals/corr/AroraLLMR15_dimension}. The next reason is strongly practical---we assume that the software developers do not want to train their models, and they pick one of the pretrained sets of embedding vectors regardless of their dimension. To understand the impact of higher dimensionality, we also conducted an experiment comparing fastText and Sent2Vec using the same number of dimensions for the proper scientific comparison. We used 300 dimensions. We pretrained both models with comparable options on the same Twitter dataset\footnote{https://archive.org/details/twitter\_cikm\_2010. We used https://pypi.org/project/tweet-preprocessor/ for preprocessing}.

We decided to use the average of embeddings for all words in the sentence to obtain a vector representation of the sentence. We used the same sentence distributions for both unsupervised and supervised tests.

\begin{table}
 \caption{F1 score comparison of embedding's robustness towards the noise.}
 \resizebox{\columnwidth}{!}{%
\begin{tabular}{ll|cccccccc} \hline
             &                  & \multicolumn{3}{c}{\textbf{Logistic Regression}}  \\
\textbf{}    & \textbf{Corpus}  & \textbf{Sent2Vec}  & \textbf{Sentence-BERT} & \textbf{USE} \\ \hline
Original        & \textbf{Ask}      & \textbf{0,908}                   & 0,880                                   & \textbf{0,908}                                \\
             & \textbf{Chat}     & \textbf{0,971}                  & \textbf{0,971}                          & 0,943                                         \\
             & \textbf{Web}      & \textbf{0,847}                    & 0,779                                   & \textbf{0,847}                                \\

\hline
Typos        & \textbf{Ask}      & 0,880                   & 0,880                                   & \textbf{0,908}                                \\
             & \textbf{Chat}     & 0,943                  & \textbf{0,990}                          & 0,962                                         \\
             & \textbf{Web}      & 0,813                    & 0,762                                   & \textbf{0,830}                                \\ \hline
Typos 50\%      & \textbf{Ask}      & 0,853                   & 0,862                                   & \textbf{0,899}                                \\
             & \textbf{Chat}     & \textbf{0,933}                   & 0,924                                   & \textbf{0,933}                                \\
             & \textbf{Web}    & 0,779                  & 0,779                                   & \textbf{0,796}                                \\ \hline
Filler Words & \textbf{Ask}   & 0,871                    & \textbf{0,880}                          & \textbf{0,880}                                \\
             & \textbf{Chat}  & 0,933                     & 0,905                                   & \textbf{0,943}                                \\
             & \textbf{Web}  & \textbf{0,779}                   & 0,711                                   & 0,762         \\ \hline                               
\end{tabular}%
}
\label{tab:noise}
\end{table}

% Please add the following required packages to your document preamble:
% \usepackage{graphicx}
\begin{table*}
\centering
\caption{F1 score comparison of public NLU services and open-source algorithms for intent recognition.}
\label{tab:results}
\resizebox{2\columnwidth}{!}{%
\begin{tabular}{ll|cccc|ccc|ccc|ccc}
\hline
                 &                            & \multicolumn{1}{l}{} & \multicolumn{1}{l}{} & \multicolumn{1}{l}{} & \multicolumn{1}{l|}{} & \multicolumn{3}{c|}{\textbf{Cosine Similarity}}                                                     & \multicolumn{3}{c|}{\textbf{SVM}}                                                                                                            & \multicolumn{3}{c}{\textbf{Logistic Regression}}                                                                                            \\
                 & \textbf{Corpus}            & \textbf{Dialogflow}  & \textbf{Luis}        & \textbf{Watson}      & \textbf{Rasa}         & \textbf{Sent2Vec} & \textbf{\begin{tabular}[c]{@{}c@{}}Sentence\\ BERT\end{tabular}} & \textbf{USE} & \multicolumn{1}{l}{\textbf{Sent2Vec}} & \textbf{\begin{tabular}[c]{@{}c@{}}Sentence\\ BERT\end{tabular}} & \multicolumn{1}{l|}{\textbf{USE}} & \multicolumn{1}{l}{\textbf{Sent2Vec}} & \textbf{\begin{tabular}[c]{@{}c@{}}Sentence\\ BERT\end{tabular}} & \multicolumn{1}{l}{\textbf{USE}} \\ \hline
\textbf{Orignal} & \textbf{Ask}               & 0,807                & 0,880                & 0,899                & 0,889                 & 0,807             & 0,834                                                            & 0,871        & 0,899                                 & 0,889                                                            & 0,871                             & \textbf{0,908}                        & 0,880                                                            & \textbf{0,908}                   \\
\textbf{}        & \textbf{Chat}              & 0,877                & 0,952                & 0,915                & \textbf{0,981}        & 0,933             & 0,933                                                            & 0,933        & \textbf{0,981}                        & 0,952                                                            & 0,943                             & 0,971                                 & 0,971                                                            & 0,943                            \\
\textbf{}        & \textbf{Web}               & 0,830                & 0,779                & 0,779                & 0,813                 & 0,745             & 0,627                                                            & 0,762        & 0,627                                 & 0,644                                                            & 0,661                             & \textbf{0,847}                        & 0,779                                                            & \textbf{0,847}                   \\ \hline
\textbf{Enrich}  & \textbf{Ask}               & 0,816                & 0,871                & \textbf{0,908}       & 0,899                 & 0,761             & 0,798                                                            & 0,834        & 0,871                                 & 0,871                                                            & 0,889                             & 0,889                                 & 0,862                                                            & 0,899                            \\
\textbf{}        & \textbf{Chat}              & 0,886                & 0,943                & 0,924                & 0,933                 & 0,915             & 0,943                                                            & 0,924        & 0,962                                 & 0,962                                                            & 0,924                             & 0,952                                 & 0,943                                                            & \textbf{0,971}                   \\
\textbf{}        & \textbf{Web}               & 0,762                & 0,745                & 0,745                & 0,711                 & 0,644             & 0,627                                                            & 0,779        & 0,610                                 & 0,644                                                            & 0,661                             & 0,762                                 & 0,813                                                            & \textbf{0,847}                   \\ \hline
\textbf{CIIRC}   & \textbf{CIIRC Small}       & 0,645                & 0,750                & 0,772                & 0,700                 & 0,891             & 0,922                                                            & 0,936        & 0,845                                 & 0,881                                                            & 0,881                             & 0,895                                 & 0,922                                                            & \textbf{0,940}                   \\
\textbf{}        & \textbf{CIIRC Small - HF}  & 0,586                & 0,690                & 0,704                & 0,654                 & 0,673             & 0,822                                                            & 0,704        & 0,754                                 & 0,809                                                            & 0,786                             & 0,850                                 & \textbf{0,859}                                                   & 0,836                            \\
\textbf{}        & \textbf{CIIRC Small - HFR} & 0,568                & 0,618                & 0,718                & 0,622                 & 0,664             & 0,768                                                            & 0,686        & 0,695                                 & 0,795                                                            & 0,695                             & 0,818                                 & 0,804                                                            & \textbf{0,822}                   \\
\textbf{}        & \textbf{CIIRC - HFR}       & 0,904                & 0,916                & 0,934                & 0,937                 & 0,802             & 0,971                                                            & 0,969        & 0,912                                 & 0,957                                                            & 0,966                             & 0,929                                 & \textbf{0,975}                                                   & 0,969                            \\ \hline
\end{tabular}%
}
\end{table*}

% Please add the following required packages to your document preamble:
% \usepackage{graphicx}
\begin{table*}
\centering
\caption{Comparison of required computational requirements and price per request.}
\label{tab:parameters}
\resizebox{2\columnwidth}{!}{%
\begin{tabular}{l|cccc|ccc|ccc|ccc}
\hline
                                 & \multicolumn{1}{l}{} & \multicolumn{1}{l}{} & \multicolumn{1}{l}{} & \multicolumn{1}{l|}{} & \multicolumn{3}{c|}{\textbf{Cosine Similarity}}                                                     & \multicolumn{3}{c|}{\textbf{SVM}}                                                                   & \multicolumn{3}{c}{\textbf{Logistic Regression}}                                                    \\
                                 & \textbf{DialogFlow}  & \textbf{Luis}        & \textbf{Watson}      & \textbf{Rasa}         & \textbf{Sent2Vec} & \textbf{\begin{tabular}[c]{@{}c@{}}Sentence\\ BERT\end{tabular}} & \textbf{USE} & \textbf{Sent2Vec} & \textbf{\begin{tabular}[c]{@{}c@{}}Sentence\\ BERT\end{tabular}} & \textbf{USE} & \textbf{Sent2Vec}  & \textbf{\begin{tabular}[c]{@{}c@{}}Sentence\\ BERT\end{tabular}} & \textbf{USE} \\ \hline
\textbf{Queries Per Hour}        & -                    & -                    & -                    & 113 467                & 2 121 428         & 79 464                                                           & 2 475 000    & 4 647 887         & 77 604                                                           & 2 001 010    & \textbf{8 658 892} & 78 893                                                           & 2 790 697    \\
\textbf{RAM (MB)}                & -                    & -                    & -                    & 358                   & \textbf{120}      & 738                                                              & 1677         & \textbf{120}      & 738                                                              & 1747         & \textbf{120}       & 738                                                              & 1650         \\
\textbf{CPU Cores}               & -                    & -                    & -                    & 4                     & \textbf{2}        & 8                                                                & 6            & \textbf{2}        & 8                                                                & 6            & \textbf{2}         & 8                                                                & 6            \\
\textbf{EC2 Instance Type}       & -                    & -                    & -                    & a1.xlarge             & t4g.micro         & a1.2xlarge                                                       & a1.xlarge    & t4g.micro         & a1.2xlarge                                                       & a1.xlarge    & t4g.micro          & a1.2xlarge                                                       & a1.xlarge    \\
\textbf{Price Per Request (USD)} & 2,00E-3              & 1,50E-3             & 3,50E-3              & 8,99E-7               & 3,96E-9           & 2,57E-6                                                          & 8,99E-7      & 1,81E-9           & 2,63E-6                                                          & 5,10E-8      & \textbf{9,70E-10}  & 2,59E-6                                                          & 3,66E-8      \\ \hline
\end{tabular}%
}
\end{table*}

% Please add the following required packages to your document preamble:
% \usepackage{graphicx}
\begin{table}
 \caption{F1 score comparison of public NLU services for intent recognition between the years 2018 and 2020.}
\resizebox{\columnwidth}{!}{%
\begin{tabular}{c|cc|cc|cc|cc}
\hline
              & \multicolumn{2}{c|}{\textbf{Dialogflow}}                                                                                       & \multicolumn{2}{c|}{\textbf{Luis}}                                                                                            & \multicolumn{2}{c|}{\textbf{Watson}}                                                                                          & \multicolumn{2}{c}{\textbf{Rasa}}                                                                                            \\
              \textbf{Corpus} & \textbf{2018} & \textbf{2020}  &   \textbf{2018} & \textbf{2020} &   \textbf{2018} & \textbf{2020}  &   \textbf{2018} & \textbf{2020}   \\ \hline
\textbf{Ask}  & 0,782         & 0,807                                                                                                & 0,742         & 0,880                                                                                               & 0,818         & \textbf{0,899}                                                                                       & 0,708         & 0,889                                                                                               \\
\textbf{Chat} & 0,935         & 0,877                                                                                                & 0,978         & 0,952                                                                                                & 0,968         & 0,915                                                                                                & 0,978         & \textbf{0,981}                                                                                       \\
\textbf{Web}  & 0,627         & \textbf{0,830}                                                                                       & 0,690         & 0,779                                                                                               & 0,630         & 0,779                                                                                               & 0,493         & 0,813                                                                                              \\ \hline
\end{tabular}%
}
\label{tab:improvent_remote}
\end{table}

We show the results in Table \ref{tab:embeddings}. Sent2Vec delivers better results in the majority of cases. FastText is better in a few cases, where it can utilize the n-gram information on unknown words. We continued to use Sent2Vec embeddings in the experiments that follow, thanks to those results.

Table \ref{tab:results} shows the F1 score comparison of public NLU services and open-source algorithms for intent recognition. We can notice several outcomes. The logistic regression is the best classifier for all sentence embedding approaches. Universal Sentence Encoder is the best sentence embedding in the majority of cases. Moreover, the open-source algorithms achieve comparable or better results than public NLU services in a remarkable number of cases. 

To evaluate the intent recognizers' noise robustness, we measured their F1 scores on datasets with added header and footer phrases, filler words, and typos. For the sake of simplicity, we have used only the best performing combinations; see Table \ref{tab:noise}. We can see that all types of noise decrease the performance of Sent2Vec. Sentence-BERT and Universal Sentence Encoder are robust to noise in most cases, except for adding filler words to Chatbot and Web Applications corpora.

We also compared the impact of noise on public NLU services and open-source algorithms. We present the comparison in Table \ref{tab:results}. The comparison shows that public NLU services are mostly robust to noise. The only exception is the Web Application corpus. However, the best results overall are achieved by logistic regression paired with Universal Sentence Encoder. The worst robustness has cosine similarity. 

We measured the computational requirements of the intent recognizers and their cost per one request. Both of those properties are closely linked. Table \ref{tab:parameters} presents the results. Measuring speed, RAM, and CPU cores do not make sense for public NLU services because those properties and requirements are handled by public NLU services' providers and are not observable. Thus, we conducted those measurements for open-source algorithms only. We can see that the Sent2Vec with logistic regression is the fastest algorithm with the lowest computational demands. It is followed by Universal Sentence Encoder, Sentence-BERT, and Rasa. Universal Sentence Encoder is the fastest of those three. However, it has the highest memory requirements. Rasa and Sentence-BERT are slower. However, they need a smaller amount of memory.

The last property we used for evaluation was cost. We present the cost of the public NLU services and open-source algorithms in Table \ref{tab:parameters}. The results clearly show that open-source algorithms are cheaper than public NLU services by several orders of magnitude.

Lastly, we compare the current results of public NLU services with the results presented in \cite{Braun2018} from 2018 to see if they are getting better over time. We show the improvements of public NLU services over two years in Table~\ref{tab:improvent_remote}.

\section{Conclusion}
We compared public NLU services with open-source algorithms for intent recognition. We selected the most popular public NLU services and open-source algorithms for comparison based on the availability of pretrained models, implementation simplicity, and proven accuracy. The evaluation criteria were classification accuracy, robustness to noise, computational demand, and cost.

We published a novel CIIRC dataset. It is a dataset for intent recognition that contains types of noise typical for real-world scenarios. We used CIIRC dataset to evaluate the intent recognition algorithm's robustness to noise.

The final result of the comparison depends on the application in which the intent recognition algorithm will be used and developer preferences. We suggest several options:

The logistic regression paired with Universal Sentence Encoder achieves the best accuracy. Although, even simpler approaches such as Sent2Vec and logistic regression achieve better classification accuracy than public NLU services in a surprising number of cases.

Experiments showed that public NLU services possess a high degree of robustness to noise. However, the best results were achieved by Universal Sentence Encoder with logistic regression.

The big advantage of public NLU services is the fact that developers do not have to worry about computational requirements. The public NLU services' providers manage the operation and scaling of those services for runtime and training. For open-source algorithms, approaches that use Sent2Vec embeddings are the fastest and least computationally demanding.

Our evaluation shows that open-source algorithms are several orders of magnitude cheaper than public NLU services based on the calculated cost per request in USD.

Hence, from our results, we can conclude the following. For developers who do not want to worry about operations and scaling, we recommend using one of the public NLU services. We recommend using Universal Sentence Encoder and logistic regression for developers who want the highest classification accuracy and high robustness to noise. And lastly, for developers who prioritize low cost and low computational demand, we recommend using Sent2Vec and logistic regression.

\section*{Acknowledgments}
%The acknowledgments should go immediately before the reference. Do not number the acknowledgments section.
%Do not include this section when submitting your paper for review.
This research has been partially supported by the Grant Agency of the Czech Technical University in Prague, grant No. SGS20/092/OHK3/1T/37.

% \bibliographystyle{acl_natbib}
% BibTeX users please use one of
% \bibliographystyle{spbasic}  
    % basic style, author-year citations
\bibliographystyle{spmpsci}      % mathematics and physical sciences
%\bibliographystyle{spphys}       % APS-like style for physics
%\bibliography{}   % name your BibTeX data base
\bibliography{references}

\begin{thebibliography}{10}
\providecommand{\url}[1]{{#1}}
\providecommand{\urlprefix}{URL }
\expandafter\ifx\csname urlstyle\endcsname\relax
  \providecommand{\doi}[1]{DOI~\discretionary{}{}{}#1}\else
  \providecommand{\doi}{DOI~\discretionary{}{}{}\begingroup
  \urlstyle{rm}\Url}\fi

\bibitem{DBLP:journals/corr/AroraLLMR15_dimension}
Arora, S., Li, Y., Liang, Y., Ma, T., Risteski, A.: Random walks on context
  spaces: Towards an explanation of the mysteries of semantic word embeddings.
\newblock CoRR \textbf{abs/1502.03520} (2015).
\newblock \urlprefix\url{http://arxiv.org/abs/1502.03520}

\bibitem{barros2019unsupervised}
Barros, J.M., Buitelaar, P., Duggan, J., Rebholz-Schuhmann, D.: Unsupervised
  classification of health content on reddit.
\newblock In: Proceedings of the 9th International Conference on Digital Public
  Health, pp. 85--89 (2019)

\bibitem{bocklisch2017rasa}
Bocklisch, T., Faulkner, J., Pawlowski, N., Nichol, A.: Rasa: Open source
  language understanding and dialogue management.
\newblock arXiv preprint arXiv:1712.05181  (2017)

\bibitem{DBLP:journals/corr/BojanowskiGJM16_fasttext}
Bojanowski, P., Grave, E., Joulin, A., Mikolov, T.: Enriching word vectors with
  subword information.
\newblock CoRR \textbf{abs/1607.04606} (2016).
\newblock \urlprefix\url{http://arxiv.org/abs/1607.04606}

\bibitem{Braun2018}
Braun, D., Hernandez-Mendez, A., Matthes, F., Langen, M.: {Evaluating Natural
  Language Understanding Services for Conversational Question Answering
  Systems} (August), 174--185 (2018).
\newblock \doi{10.18653/v1/w17-5522}

\bibitem{Brich2018}
Brich, T.: {Semantic Sentence Similarity for Intent Recognition Task} (May)
  (2018)

\bibitem{bunk2020diet}
Bunk, T., Varshneya, D., Vlasov, V., Nichol, A.: Diet: Lightweight language
  understanding for dialogue systems.
\newblock arXiv preprint arXiv:2004.09936  (2020)

\bibitem{Cer2018}
Cer, D., Yang, Y., yi~Kong, S., Hua, N., Limtiaco, N., {St. John}, R.,
  Constant, N., Guajardo-C{\'{e}}spedes, M., Yuan, S., Tar, C., Sung, Y.H.,
  Strope, B., Kurzweil, R.: {Universal sentence encoder for English}.
\newblock EMNLP 2018 - Conference on Empirical Methods in Natural Language
  Processing: System Demonstrations, Proceedings pp. 169--174 (2018).
\newblock \doi{10.18653/v1/d18-2029}

\bibitem{cortes1995support}
Cortes, C., Vapnik, V.: Support-vector networks.
\newblock Machine learning \textbf{20}(3), 273--297 (1995)

\bibitem{dai2017social}
Dai, X., Bikdash, M., Meyer, B.: From social media to public health
  surveillance: Word embedding based clustering method for twitter
  classification.
\newblock In: SoutheastCon 2017, pp. 1--7. IEEE (2017)

\bibitem{BERT2019}
Devlin, J., Chang, M.W., Lee, K., Toutanova, K.: {BERT: Pre-training of deep
  bidirectional transformers for language understanding}.
\newblock NAACL HLT 2019 - 2019 Conference of the North American Chapter of the
  Association for Computational Linguistics: Human Language Technologies -
  Proceedings of the Conference \textbf{1}(Mlm), 4171--4186 (2019)

\bibitem{Edizel2019_misspeling}
Edizel, B., Piktus, A., Bojanowski, P., Ferreira, R., Grave, E., Silvestri, F.:
  {Misspelling oblivious word embeddings}.
\newblock NAACL HLT 2019 - 2019 Conference of the North American Chapter of the
  Association for Computational Linguistics: Human Language Technologies -
  Proceedings of the Conference \textbf{1}, 3226--3234 (2019).
\newblock \doi{10.18653/v1/n19-1326}

\bibitem{focil2017tweets}
F{\'o}cil-Arias, C., Ziiniga, J., Sidorov, G., Batyrshin, I., Gelbukh, A.: A
  tweets classifier based on cosine similarity.
\newblock In: Working notes of CLEF 2017—Conference and Labs of the
  Evaluation Forum, Dublin, Ireland, pp. 11--14 (2017)

\bibitem{gabriel2020further}
Gabriel, R., Liu, Y., Gottardi, A., Eric, M., Khatri, A., Chadha, A., Chen, Q.,
  Hedayatnia, B., Rajan, P., et~al.: Further advances in open domain dialog
  systems in the third alexa prize socialbot grand challenge.
\newblock Proc. Alexa Prize  (2020)

\bibitem{Hagiwara2019_github_typos}
Hagiwara, M., Mita, M.: {GitHub Typo Corpus: A Large-Scale Multilingual Dataset
  of Misspellings and Grammatical Errors}  (2019).
\newblock \urlprefix\url{http://arxiv.org/abs/1911.12893}

\bibitem{hochreiter1997long_LSTM}
Hochreiter, S., Schmidhuber, J.: Long short-term memory.
\newblock Neural computation \textbf{9}(8), 1735--1780 (1997)

\bibitem{fastextZip}
Joulin, A., Grave, E., Bojanowski, P., Douze, M., J{\'{e}}gou, H., Mikolov, T.:
  Fasttext.zip: Compressing text classification models.
\newblock CoRR \textbf{abs/1612.03651} (2016).
\newblock \urlprefix\url{http://arxiv.org/abs/1612.03651}

\bibitem{khatri2018alexa}
Khatri, C., Venkatesh, A., Hedayatnia, B., Gabriel, R., Ram, A., Prasad, R.:
  Alexa prize—state of the art in conversational ai.
\newblock AI Magazine \textbf{39}(3), 40--55 (2018)

\bibitem{DBLP:journals/corr/LiangLSBLS17_GAN_fooled_neural_network}
Liang, B., Li, H., Su, M., Bian, P., Li, X., Shi, W.: Deep text classification
  can be fooled.
\newblock CoRR \textbf{abs/1704.08006} (2017).
\newblock \urlprefix\url{http://arxiv.org/abs/1704.08006}

\bibitem{FastBert2020}
Liu, W., Zhou, P., Zhao, Z., Wang, Z., Deng, H., Ju, Q.: {FastBERT: a
  Self-distilling BERT with Adaptive Inference Time}  (2020).
\newblock \urlprefix\url{http://arxiv.org/abs/2004.02178}

\bibitem{Liu2019_benchmark_NLU}
Liu, X., Eshghi, A., Swietojanski, P., Rieser, V.: {Benchmarking Natural
  Language Understanding Services for building Conversational Agents}  (2019).
\newblock \urlprefix\url{http://arxiv.org/abs/1903.05566}

\bibitem{mccullagh2018generalized}
McCullagh, P.: Generalized linear models.
\newblock Routledge (2018)

\bibitem{DBLP:journals/corr/MikolovSCCD13_word2vec}
Mikolov, T., Sutskever, I., Chen, K., Corrado, G., Dean, J.: Distributed
  representations of words and phrases and their compositionality.
\newblock CoRR \textbf{abs/1310.4546} (2013).
\newblock \urlprefix\url{http://arxiv.org/abs/1310.4546}

\bibitem{macro_f1_score}
Opitz, J., Burst, S.: Macro f1 and macro f1  (2019)

\bibitem{pgj2017sent2vec}
Pagliardini, M., Gupta, P., Jaggi, M.: {Unsupervised Learning of Sentence
  Embeddings using Compositional n-Gram Features}.
\newblock In: NAACL 2018 - Conference of the North American Chapter of the
  Association for Computational Linguistics (2018)

\bibitem{scikit-learn}
Pedregosa, F., Varoquaux, G., Gramfort, A., Michel, V., Thirion, B., Grisel,
  O., Blondel, M., Prettenhofer, P., Weiss, R., Dubourg, V., Vanderplas, J.,
  Passos, A., Cournapeau, D., Brucher, M., Perrot, M., Duchesnay, E.:
  Scikit-learn: Machine learning in {P}ython.
\newblock Journal of Machine Learning Research \textbf{12}, 2825--2830 (2011)

\bibitem{pichl2020alquist}
Pichl, J., Marek, P., Konr{\'a}d, J., Lorenc, P., Duy, T.V., {\v{S}}ediv{\'y},
  J.: Alquist 3.0: Alexa prize bot using conversational knowledge graph.
\newblock Alexa Prize Socialbot Grand Challenge 3 Proceedings  (2020)

\bibitem{pichl2018alquist}
Pichl, J., Marek, P., Konr{\'a}d, J., Matul{\'\i}k, M., Nguyen, H.L.,
  {\v{S}}ediv{\'y}, J.: Alquist: The alexa prize socialbot.
\newblock arXiv preprint arXiv:1804.06705  (2018)

\bibitem{Pichl2018}
Pichl, J., Marek, P., Matul{\'{i}}k, M., {\v{S}}ediv{\'{y}}, J.: {Alquist 2.0:
  Alexa Prize Socialbot Based on Sub-Dialogue Models} (Section 2) (2018).
\newblock \urlprefix\url{https://aws.amazon.com/codepipeline}

\bibitem{ram2018conversational}
Ram, A., Prasad, R., Khatri, C., Venkatesh, A., Gabriel, R., Liu, Q., Nunn, J.,
  Hedayatnia, B., Cheng, M., Nagar, A., et~al.: Conversational ai: The science
  behind the alexa prize.
\newblock arXiv preprint arXiv:1801.03604  (2018)

\bibitem{Reimers2019}
Reimers, N., Gurevych, I.: {Sentence-BERT: Sentence Embeddings using Siamese
  BERT-Networks} pp. 3980--3990 (2019).
\newblock \doi{10.18653/v1/d19-1410}

\bibitem{Salvaris_2018_GAN}
Salvaris, M., Dean, D., Tok, W.H.: Generative adversarial networks.
\newblock Deep Learning with Azure p. 187–208 (2018)

\bibitem{Shivakumar2019}
Shivakumar, P.G., Li, H., Knight, K., Georgiou, P.: Learning from past
  mistakes: Improving automatic speech recognition output via noisy-clean
  phrase context modeling.
\newblock APSIPA Transactions on Signal and Information Processing \textbf{8},
  1--16 (2019).
\newblock \doi{10.1017/ATSIP.2018.31}

\bibitem{Sun2020_typos_bert}
Sun, L., Hashimoto, K., Yin, W., Asai, A., Li, J., Yu, P., Xiong, C.:
  {Adv-BERT: BERT is not robust on misspellings! Generating nature adversarial
  samples on BERT}  (2020).
\newblock \urlprefix\url{http://arxiv.org/abs/2003.04985}

\bibitem{attentionis2017}
Vaswani, A., Shazeer, N., Parmar, N., Uszkoreit, J., Jones, L., Gomez, A.N.,
  Kaiser, {\L}., Polosukhin, I.: {Attention is all you need}.
\newblock Advances in Neural Information Processing Systems
  \textbf{2017-December}(Nips), 5999--6009 (2017)

\end{thebibliography}

% Non-BibTeX users please use
% \begin{thebibliography}{}
% %
% % and use \bibitem to create references. Consult the Instructions
% % for authors for reference list style.
% %
% \bibitem{RefJ}
% % Format for Journal Reference
% Author, Article title, Journal, Volume, page numbers (year)
% % Format for books
% \bibitem{RefB}
% Author, Book title, page numbers. Publisher, place (year)
% % etc
% \end{thebibliography}

\end{document}